%% file: main.tex
\pdfoutput=1
\documentclass{article}
\usepackage{ijcai22}

\usepackage{times}
\usepackage{soul}
\usepackage{url}
\usepackage[hidelinks]{hyperref}
\usepackage[utf8]{inputenc}
\usepackage[small]{caption}
\usepackage{graphicx}
\usepackage{amsmath}
\usepackage{amsthm}
\usepackage{booktabs}
\usepackage{algorithm}
\usepackage{algorithmic}
\usepackage{microtype}
\usepackage{amssymb}
\usepackage{multirow}
\usepackage{subcaption}

\usepackage{tikz}
\usetikzlibrary{arrows}
\usetikzlibrary{shapes}
\usetikzlibrary{patterns}

\newtheorem*{theorem*}{Theorem}

\makeatletter
\newcommand*\bigcdot{\mathpalette\bigcdot@{.5}}
\newcommand*\bigcdot@[2]{\mathbin{\vcenter{\hbox{\scalebox{#2}{$\m@th#1\bullet$}}}}}
\makeatother

\newcommand{\bx}{\boldsymbol{x}}
\newcommand{\bz}{\boldsymbol{z}}
\newcommand{\br}{\boldsymbol{r}}
\newcommand{\bt}{\boldsymbol{\theta}}
\newcommand{\cL}{\mathcal{L}}
\newcommand{\cI}{\mathcal{I}}
\newcommand{\cA}{\mathcal{A}}
\newcommand{\cP}{\mathcal{P}}
\newcommand{\cK}{\mathcal{K}}
\newcommand{\cX}{\mathcal{X}}
\newcommand{\cY}{\mathcal{Y}}

\title{\mbox{Dual Contrastive Learning: Text Classification via Label-Aware Data Augmentation}}

\author{
Qianben Chen\textsuperscript{\rm 1,2}\and
Richong Zhang\textsuperscript{\rm 1,2}\footnote{Corresponding author:\texttt{zhangrc@act.buaa.edu.cn}}\and
Yaowei Zheng\textsuperscript{\rm 1,2}\And
Yongyi Mao\textsuperscript{\rm 3}\\
\affiliations
\textsuperscript{\rm 1}SKLSDE, School of Computer Science and Engineering, Beihang University, Beijing, China\\
\textsuperscript{\rm 2}Beijing Advanced Institution on Big Data and Brain Computing, Beihang University, Beijing, China\\
\textsuperscript{\rm 3}School of Electrical Engineering and Computer Science, University of Ottawa, Ottawa, Canada\\
\emails
anthonychen@buaa.edu.cn,
zhangrc@act.buaa.edu.cn,
hiyouga@buaa.edu.cn,
ymao@uottawa.ca
}

\begin{document}

\maketitle

\input{abstract}
\input{intro}
\input{pre}
\input{model}
\input{exp}
\input{relworks}
\input{conclusion}

\bibliographystyle{named}
\bibliography{main}

\input{appendix}

\end{document}

%% file: abstract.tex
\begin{abstract}

Contrastive learning has achieved remarkable success in representation learning via self-supervision in unsupervised settings. However, effectively adapting contrastive learning to supervised learning tasks remains as a challenge in practice. In this work, we introduce a dual contrastive learning (DualCL) framework that simultaneously learns the features of input samples and the parameters of classifiers in the same space. Specifically, DualCL regards the parameters of the classifiers as augmented samples associating to different labels and then exploits the contrastive learning between the input samples and the augmented samples. Empirical studies on five benchmark text classification datasets and their low-resource version demonstrate the improvement in classification accuracy and confirm the capability of learning discriminative representations of DualCL. 

\end{abstract}

%% file: intro.tex
\section{Introduction}

Representation learning is at the heart of modern deep learning. In the context of unsupervised learning, contrastive learning \cite{hadsell2006dimensionality} has been recently demonstrated as an effective approach to obtain generic representations for downstream tasks \cite{he2020moco,chen2020simclr}. Briefly, unsupervised contrastive learning adopts a loss function that forces representations of different ``views'' of the same example to be similar and representations of different examples to be distinct. Recently the effectiveness of contrastive learning is justified in terms of simultaneously achieving both ``alignment'' and ``uniformity'' \cite{wang2020understanding}.

Such a contrastive learning approach has also been adapted to supervised representation learning \cite{khosla2020supervised_contrastive}, in which a similar contrastive loss is used which insists the representations of examples in the same class to be similar and those for different classes to be distinct. However, despite its demonstrated successes, such an approach appears much less principled, compared with unsupervised contrastive learning. For example, the uniformity of the representations is no longer valid; neither is it required. In fact, we argue that the standard supervised contrastive learning approach is not natural for supervised representation learning. This is at least manifested by the fact that the outcome of this approach does not give us a classifier directly and one is required to develop another classification algorithm to solve the classification task.

\begin{table}[t]
    \centering
    \resizebox{.85\columnwidth}{!}{
    \begin{tabular}{c|c}
        \resizebox{.49\columnwidth}{!}{\input{tikzfigs/concept1}} & \resizebox{.49\columnwidth}{!}{\input{tikzfigs/concept2}} \\
        \hline
        \resizebox{.49\columnwidth}{!}{\input{tikzfigs/concept3}} & \resizebox{.49\columnwidth}{!}{\input{tikzfigs/concept4}}
    \end{tabular}%
    }
    \caption{The concept of our proposed dual contrastive learning. Each color represent one class. The circle and triangle denote the representation of the input sample and the classifier respectively.}
    \label{fig:concept}
\end{table}

This paper aims at developing a more natural approach to contrastive learning in the supervised setting. A key insight in our development is that supervised representation learning ought to include learning two kinds of quantities: one is the feature $\bz$ of the input $\bx$ in an appropriate space that is sufficiently discriminative for the classification task, and the other is a classifier on that space, or alternatively the parameter $\bt$ of the classifier acting on that space; we will refer to this classifier as the ``one-example'' classifier for example $\bx$. In this view, it is natural to associate with each example $\bx$ two quantities, a vector $\bz\in\mathbb{R}^d$ for an appropriate feature space dimension $d$ and a matrix $\bt\in\mathbb{R}^{d\times K}$ defining a linear classifier for $x$ (assuming a $K$-class classification problem), note that both $\bz$ and $\bt$ depend on $x$. The representation learning problem in the supervised setting can be regarded as {\em learning to generate the pair $(\bz,\bt)$ for an input example $\bx$}. 

For the classifier $\bt$ to be valid for feature $\bz$, we only need to align the softmax transform of $\bt^T\bz$ with the label of $\bx$ using the standard cross-entropy loss. In addition, a contrastive learning approach can be used to force constraints on these $(\bz,\bt)$ representations across examples. Specifically, let $\bt^*$ denote the column of $\bt$ corresponding to the ground-truth label of $\bx$, we may design two contrastive losses. The first loss contrasts $(\bz,\bt^*)$ with many $(\bz',\bt^*)$s, where $\bz'$ is the feature of an example having a different label as $\bx$; the second contrasts $(\bz,\bt^*)$ with many $(\bz,\bt'^*)$s, where $\bt'$ is the classifier associate with an example from a different class. We refer to this learning framework as dual contrastive learning (DualCL).

Despite that we propose dual contrastive learning based on a conceptual insight, we argue that it is possible to also interpret such a learning scheme as exploiting a unique data augmentation approach. In particular, for each example $\bx$, each column of its $\bt$ can be regarded as an ``label-aware input representation'', or an augmented view of $\bx$ in the feature space with the label information infused. The figures in Table~\ref{fig:concept} illustrate the benefit of this approach, from the two figures on the left side, it can be seen that standard contrastive learning cannot make use of the label information. On the contrary, from the two figures on the right side, DualCL effectively leverages the label information to classify the input samples in their classes.

In experiments, we validate the effectiveness of DualCL on five benchmark text classification datasets. By finetuning the pretrained language model (BERT and RoBERTa) using the dual contrastive loss, DualCL achieves the best performance compared to existing supervised baselines with contrastive learning. We also find that DualCL improves the classification accuracy especially on low-resource scenarios. Furthermore, we give some explanations for DualCL through visualizing the learned representations and attention maps. 

Our contributions can be summarized as follows.

1) We propose the dual contrastive learning (DualCL) for naturally adapting the contrastive loss to supervised settings.

2) We introduce the label-aware data augmentation to obtain multiple views of input samples for the training of DualCL.

3) We empirically verify the effectiveness of the DualCL framework on five benchmark text classification datasets.

%% file: tikzfigs/concept1.tex
\begin{tikzpicture}
    
    \draw[use as bounding box,draw=white] (-3,-2) rectangle (3,1.9);
    
    \tikzstyle{textA}=[circle,draw=blue!80,fill=blue!50,inner sep=4pt,outer sep=2pt,thick]
    \tikzstyle{textB}=[circle,draw=red!80,fill=red!50,inner sep=4pt,outer sep=2pt,thick]
    \tikzstyle{repel}=[angle 60-angle 60,line width=1]
    \tikzstyle{attract}=[angle 60 reversed-angle 60 reversed,line width=1]
    \tikzstyle{same}=[-,dashed,line width=1]
    
    \node at (-3.2, 1.7) [anchor=west] (Caption) {Standard Contrastive Learning (before)};
    
    \node at ( 0.7, 1.1) [textA] (textA1) {};
    \node at (-1.1,-0.8) [textA] (textA2) {};
    \node at (-1.3, 0.9) [textA] (textA3) {};
    \node at ( 0.4,-0.3) [textA] (textA4) {};
    \node at ( 1.3,-0.8) [textA] (textA5) {};
    
    \node at (-1.8, 0.1) [textB] (textB1) {};
    \node at (-0.3, 0.3) [textB] (textB2) {};
    \node at ( 1.8, 0.9) [textB] (textB3) {};
    \node at (-1.4,-1.6) [textB] (textB4) {};
    \node at ( 0.3,-1.2) [textB] (textB5) {};
    
    \draw [attract] (textA1) to [bend left] (textA4);
    \draw [repel] (textA1) to [bend right] (textB2);
    
\end{tikzpicture}

%% file: tikzfigs/concept2.tex
\begin{tikzpicture}
    
    \draw[use as bounding box,draw=white] (-3,-2) rectangle (3,1.9);
    
    \tikzstyle{textA}=[circle,draw=blue!80,fill=blue!50,inner sep=4pt,outer sep=2pt,thick]
    \tikzstyle{textB}=[circle,draw=red!80,fill=red!50,inner sep=4pt,outer sep=2pt,thick]
    \tikzstyle{labelA}=[regular polygon,regular polygon sides=3,draw=blue!80,fill=blue!50,inner sep=2pt,outer sep=3pt,thick]
    \tikzstyle{labelB}=[regular polygon,regular polygon sides=3,draw=red!80,fill=red!50,inner sep=2pt,outer sep=3pt,thick]
    \tikzstyle{repel}=[angle 60-angle 60,line width=1]
    \tikzstyle{attract}=[angle 60 reversed-angle 60 reversed,line width=1]
    \tikzstyle{same}=[-,dashed,line width=1]
    
    \node at (-3.2, 1.7) [anchor=west] (Caption) {Dual Contrastive Learning (before)};
    
    \node at ( 0.7, 1.1) [textA] (textA1) {};
    \node at (-2.2,-1.5) [textA] (textA2) {};
    \node at ( 2.2,-1.8) [textA] (textA3) {};
    \node at ( 1.5,-0.4) [labelA] (labelA1) {};
    \node at (-0.6,-0.9) [labelA] (labelA2) {};
    \node at ( 0.8,-1.2) [labelA] (labelA3) {};
    \node at (-1.8, 0.6) [textB] (textB1) {};
    \node at (-0.9, 0.0) [textB] (textB2) {};
    \node at ( 2.5,-0.8) [textB] (textB3) {};
    \node at (-2.5,-0.7) [labelB] (labelB1) {};
    \node at (-1.2, 1.2) [labelB] (labelB2) {};
    \node at ( 2.1, 1.0) [labelB] (labelB3) {};
    
    \draw [same] (textA1) -- (labelA1);
    \draw [same] (textA2) -- (labelA2);
    \draw [same] (textA3) -- (labelA3);
    
    \draw [attract] (textB1) -- (labelB1);
    \draw [same] (textB2) -- (labelB2);
    \draw [same] (textB3) -- (labelB3);
    
    \draw [attract] (textA1) to [bend right] (labelA3);
    \draw [repel] (textA1) to [bend left] (labelB2);
    
\end{tikzpicture}

%% file: tikzfigs/concept3.tex
\begin{tikzpicture}
    
    \draw[use as bounding box,draw=white] (-3,-1.8) rectangle (3,1.9);
    
    \tikzstyle{textA}=[circle,draw=blue!80,fill=blue!50,inner sep=4pt,outer sep=2pt,thick]
    \tikzstyle{textB}=[circle,draw=red!80,fill=red!50,inner sep=4pt,outer sep=2pt,thick]
    \tikzstyle{same}=[-,dashed,line width=1]
    
    \node at (-3.2, 1.6) [anchor=west] (Caption) {Standard Contrastive Learning (after)};
    
    \node at ( 0.8, 0.5) [textA] (textA1) {};
    \node at ( 2.0,-0.1) [textA] (textA2) {};
    \node at ( 1.6,-0.9) [textA] (textA3) {};
    \node at ( 0.2,-0.8) [textA] (textA4) {};
    \node at ( 1.0,-0.4) [textA] (textA5) {};
    
    \node at (-1.7, 0.6) [textB] (textB1) {};
    \node at (-0.5, 0.1) [textB] (textB2) {};
    \node at (-1.9,-1.0) [textB] (textB3) {};
    \node at (-2.0,-0.2) [textB] (textB4) {};
    \node at (-1.2,-0.4) [textB] (textB5) {};
    
\end{tikzpicture}

%% file: tikzfigs/concept4.tex
\begin{tikzpicture}
    
    \draw[use as bounding box,draw=white] (-3,-1.8) rectangle (3,1.9);
    
    \tikzstyle{textA}=[circle,draw=blue!80,fill=blue!50,inner sep=4pt,outer sep=2pt,thick]
    \tikzstyle{textB}=[circle,draw=red!80,fill=red!50,inner sep=4pt,outer sep=2pt,thick]
    \tikzstyle{labelA}=[regular polygon,regular polygon sides=3,draw=blue!80,fill=blue!50,inner sep=2pt,outer sep=3pt,thick]
    \tikzstyle{labelB}=[regular polygon,regular polygon sides=3,draw=red!80,fill=red!50,inner sep=2pt,outer sep=3pt,thick]
    \tikzstyle{same}=[-,dashed,line width=1]
    
    \node at (-3.2, 1.6) [anchor=west] (Caption) {Dual Contrastive Learning (after)};
    
    \node at ( 1.3, 0.4) [textA] (textA1) {};
    \node at ( 2.3,-0.2) [textA] (textA2) {};
    \node at ( 2.1,-0.8) [textA] (textA3) {};
    \node at ( 1.5,-0.4) [labelA] (labelA1) {};
    \node at ( 1.9, 0.5) [labelA] (labelA2) {};
    \node at ( 1.0,-0.7) [labelA] (labelA3) {};
    
    \node at (-1.9, 0.0) [textB] (textB1) {};
    \node at (-1.3,-0.7) [textB] (textB2) {};
    \node at (-2.1,-1.2) [textB] (textB3) {};
    \node at (-2.4,-0.6) [labelB] (labelB1) {};
    \node at (-1.2,-0.0) [labelB] (labelB2) {};
    \node at (-1.2,-1.4) [labelB] (labelB3) {};
    
    \draw [same] (textA1) -- (labelA1);
    \draw [same] (textA2) -- (labelA2);
    \draw [same] (textA3) -- (labelA3);
    
    \draw [same] (textB1) -- (labelB1);
    \draw [same] (textB2) -- (labelB2);
    \draw [same] (textB3) -- (labelB3);
    
\end{tikzpicture}

%% file: pre.tex
\section{Preliminaries}

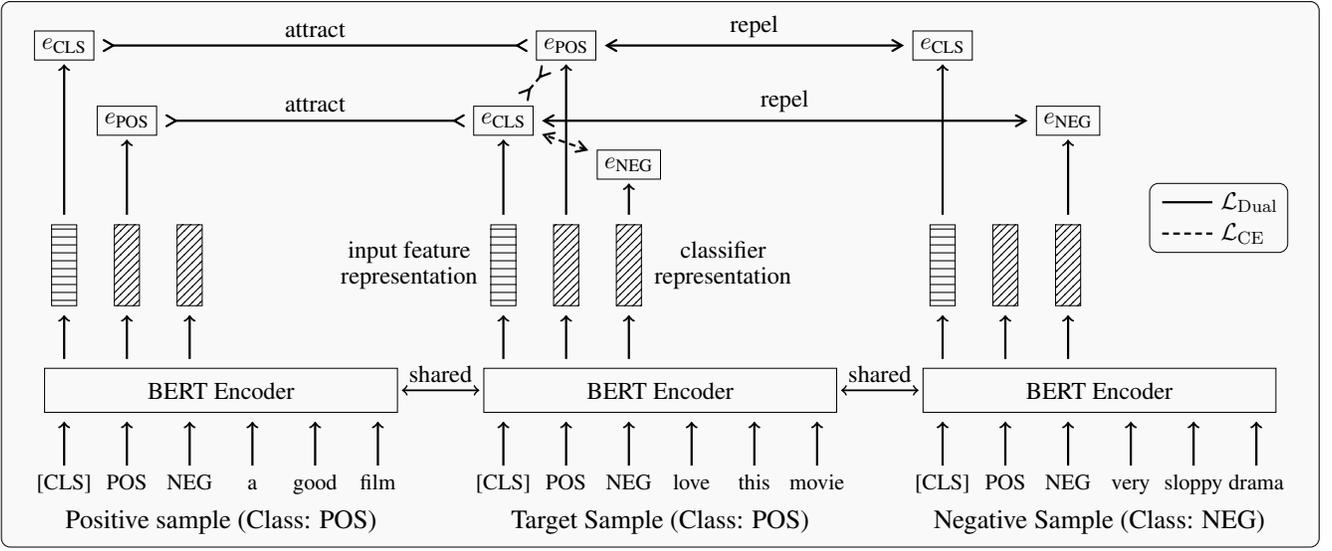
\begin{figure*}
    \centering
    \resizebox{.99\textwidth}{!}{\input{tikzfigs/model}}
    \caption{The framework of the proposed dual contrastive learning (DualCL).}
    \label{fig:model}
\end{figure*}

Consider a text classification task with $K$ classes. We assume that the given dataset $\{\bx_i,y_i\}_{i=1}^N$ contains $N$ training samples, where $\bx_i\in\mathbb{R}^L$ is the input sentence consisting of $L$ words and $y_i\in\{1,2,\cdots,K\}$ is the label assigned to the input. Throughout this study, we denote the set of indexes of the training samples by $\cI=\{1,2,\cdots,N\}$ and the set of indexes of the labels by $\cK=\{1,2,\cdots,K\}$.

Before introducing our method, we look at the family of self-supervised contrastive learning, whose effectiveness has been widely confirmed in many studies. Given $N$ training samples $\{\bx_i\}_{i=1}^N$ with a number of augmented samples, where each sample has at least one augmented sample in the dataset. Let $j(i)$ be the index of the augmented one derived from the $i$-th sample, the standard contrastive loss is defined as:
\begin{equation}
\cL_{\rm self}=\frac1N\sum_{i\in\cI}-\log\frac{\exp(\bz_i\bigcdot\bz_{j(i)}/\tau)}{\sum_{a\in\cA_i}\exp(\bz_i\bigcdot\bz_{a}/\tau)}
\end{equation}
where $\bz_i$ is the normalized representation of $\bx_i$, $\cA_i:=\cI\setminus\{i\}$ is the set of indexes of the contrastive samples, the $\bigcdot$ symbol denotes the dot product and $\tau\in\mathbb{R}^+$ is the temperature factor.

Here, we define the $i$-th sample as an {\em anchor}, the $j(i)$-th sample is a {\em positive} sample and the remaining $N-2$ samples are {\em negative} samples regarding to the $i$-th sample.

However, self-supervised contrastive learning is unable to leverage the supervised signals. Previous study \cite{khosla2020supervised_contrastive} incorporates supervision to contrastive learning in a straightforward way. It simply takes the samples from the same class as positive samples and the samples from different classes as negative samples. The following contrastive loss is defined for supervised tasks:
\begin{equation}
\cL_{\rm sup}=\frac1N\sum_{i\in\cI}\frac{1}{|\cP_i|}\sum_{p\in\cP_i}-\log\frac{\exp(\bz_i\bigcdot\bz_p/\tau)}{\sum_{a\in\cA_i}\exp(\bz_i\bigcdot\bz_{a}/\tau)}
\end{equation}
where $\cP_i:=\{p\in\cA_i:y_p=y_i\}$ is the set of indexes of positive samples, and $|\cP_i|$ is the cardinality of $\cP_i$.

Although this approach has shown its superiority, we still need to learn a linear classifier using the cross-entropy loss apart from the contrastive term. This is because the contrastive loss can only learn generic representations for the input examples. Thus, we argue that the supervised contrastive learning developed so far appears to be a naive adaptation of unsupervised contrastive learning to the classification problem. One may expect a more elegant approach for contrastive learning in supervised settings.

%% file: tikzfigs/model.tex
\begin{tikzpicture}[font=\large]
    
    \draw [use as bounding box,rounded corners,fill=gray!5] (-0.5,0.5) rectangle (20.5,9.2);
    
    \def\examplevar{1.0}
    
    \tikzstyle{BERT}=[draw=black,minimum width=160pt,minimum height=20pt,outer sep=2pt]
    \node at (3.0, 3.0) [BERT] (BERT1) {BERT Encoder};
    \node at (10.0, 3.0) [BERT] (BERT2) {BERT Encoder};
    \node at (17.0, 3.0) [BERT] (BERT3) {BERT Encoder};
    
    \draw [<->,line width=0.8] (BERT1) -- (BERT2) node [midway,above] {shared};
    \draw [<->,line width=0.8] (BERT2) -- (BERT3) node [midway,above] {shared};
    
    \node at (3.0, 0.9) [font=\Large] {Positive sample (Class: POS)};
    \node at (10.0, 0.9) [font=\Large] {Target Sample (Class: POS)};
    \node at (17.0, 0.9) [font=\Large] {Negative Sample (Class: NEG)};
    
    \foreach \x in {0,7,14} {
        \foreach \y in {0.5, 1.5, 2.5, 3.5, 4.5, 5.5} {
            \draw [->,line width=1] (\x+\y, 1.8) -- (\x+\y, 2.5);
        }
        \foreach \y in {0.5, 1.5, 2.5} {
            \draw [->,line width=1] (\x+\y, 3.5) -- (\x+\y, 4.2);
            \pgfmathparse{int((\y-0.5)*10)}
            \let\r\pgfmathresult
            \ifnum\r=0
                \draw [pattern=horizontal lines] (\x+\y-0.2, 4.35) rectangle (\x+\y+0.2, 5.64);
            \else
                \draw [pattern=north east lines] (\x+\y-0.2, 4.35) rectangle (\x+\y+0.2, 5.64);
            \fi
        }
    }

    \foreach \y/\t in {0.5/[CLS], 1.5/POS, 2.5/NEG, 3.5/a, 4.5/good, 5.5/film} {
        \node at (\y, 1.5) [font=\normalsize] {\t\strut};
    }
    \foreach \y/\t in {0.5/[CLS], 1.5/POS, 2.5/NEG, 3.5/love, 4.5/this, 5.5/movie} {
        \node at (7+\y, 1.5) [font=\normalsize] {\t\strut};
    }
    \foreach \y/\t in {0.5/[CLS], 1.5/POS, 2.5/NEG, 3.5/very, 4.5/sloppy, 5.5/drama} {
        \node at (14+\y, 1.5) [font=\normalsize] {\t\strut};
    }
    
    \node at (6.0, 5.0) [text width=80pt,text badly centered] {input feature representation};
    \node at (11.0, 5.0) [text width=80pt,text badly centered] {classifier representation};
    
    \draw [->,line width=1] (0.5, 5.8) -- (0.5, 8.2);
    \draw [->,line width=1] (1.5, 5.8) -- (1.5, 7.0);
    
    \draw [->,line width=1] (7+0.5, 5.8) -- (7+0.5, 7.0);
    \draw [->,line width=1] (7+1.5, 5.8) -- (7+1.5, 8.2);
    \draw [->,line width=1] (7+2.5, 5.8) -- (7+2.5, 6.3);
    
    \draw [->,line width=1] (14+0.5, 5.8) -- (14+0.5, 8.2);
    \draw [->,line width=1] (14+2.5, 5.8) -- (14+2.5, 7.0);

    \tikzstyle{NODE}=[draw=black,outer sep=4pt]
    
    \node at (0.5, 8.5) [NODE] (CLS1) {$e_\text{CLS}$};
    \node at (1.5, 7.3) [NODE] (LAB1) {$e_\text{POS}$};
    
    \node at (7+0.5, 7.3) [NODE] (CLS2) {$e_\text{CLS}$};
    \node at (7+1.5, 8.5) [NODE] (LAB2) {$e_\text{POS}$};
    \node at (7+2.5, 6.6) [NODE] (NEG2) {$e_\text{NEG}$};
    
    \node at (14+0.5, 8.5) [NODE] (CLS3) {$e_\text{CLS}$};
    \node at (14+2.5, 7.3) [NODE] (LAB3) {$e_\text{NEG}$};
    
    \tikzstyle{repel}=[angle 60-angle 60,line width=1]
    \tikzstyle{attract}=[angle 60 reversed-angle 60 reversed,line width=1]
    
    \draw [attract] (CLS1) -- (LAB2) node [midway,above] {attract};
    \draw [repel] (CLS3) -- (LAB2) node [midway,above] {repel};
    
    \draw [attract] (LAB1) -- (CLS2) node [midway,above] {attract};
    \draw [repel] (LAB3) -- (CLS2) node [midway,above] {repel};
    
    \draw [attract,densely dashed] (LAB2) -- (CLS2);
    \draw [repel,densely dashed] (NEG2) -- (CLS2);
    
    \draw [rounded corners] (17.8,5.2) rectangle (20.0,6.3);
    
    \draw [-,line width=1] (18.0,6.0) -- (18.8,6.0) node [anchor=west] {$\cL_{\rm Dual}$};
    \draw [-,line width=1,densely dashed] (18.0,5.5) -- (18.8,5.5) node [anchor=west] {$\cL_{\rm CE}$};

\end{tikzpicture}

%% file: model.tex
\section{Dual Contrastive Learning}

\input{mainresult}

In this paper, we propose a supervised contrastive learning approach that learns ``dual'' representations. The first one is the input representation of discriminative features for the classification task in an appropriate space, and the second one is a classifier, or equivalently the parameter of the classifier in that space. Let $\bz_i\in\mathbb{R}^d$ be the feature of an input example $\bx_i$ and $\bt_i\in\mathbb{R}^{d\times K}$ be the classifier associating to $\bx_i$. Our aim is to learn (normalized) representations of $\bz_i$ and $\bt_i$ to align the softmax transform of $\bt_i^T\bz_i$ with the label of $\bx_i$ using the proposed approach.

\subsection{Label-Aware Data Augmentation}

In order to obtain different views of the training samples, we utilize the idea of data augmentation to obtain the representations of feature $\bz_i$ and classifier $\bt_i$. We regard the $k$-th column of $\bt_i$ as a unique view of input example $\bx_i$ associated with label $k$, denoted by $\bt_i^k$. We call $\bt_i^k$ as the label-aware input representation since it is an augmented view of $\bx_i$ with the information of label $k$ infused.

Here, we introduce label-aware data augmentation. Noting that we do not actually introduce additional samples, we get the $K+1$ views for each sample just in one single feed-forward procedure. Specifically, we use a pretrained encoder to learn $1$ feature representation and $K$ label-aware input representations, where $K$ is the number of classes. Pretrained language models (PLMs) show remarkable performance in extracting natural language representations. Thus, we adopt PLMs as the pretrained encoder. Let the pretrained encoder ({\em e.g.} BERT, RoBERTa) be $f$. 
We feed both the input sentence and all the possible labels to encoder $f$ and regard each label as one token. Specifically, we list all the labels $\{1,\cdots,K\}$ and insert them before the input sentence $\bx_i$. This process forms a new sequence $\br_i\in\mathbb{R}^{L+K}$. Then an encoder $f$ is exploited to extract features of each token in this sequence.

We take the feature of the [CLS] token as the representation of each input sentence, and the feature of the token corresponding to each label as the label-aware input representation. We denote the feature representation of input $\bx_i$ by $\bz_i$ and the label-aware input representation of label $k$ by $\bt_i^k$. In practice, we take the name of labels as the tokens to form sequence $\br_i$, such as ``positive'', ``negative'', etc. For the labels containing multiple words, we take the mean-pooling of the token features to obtain the label-aware input representations.

\subsection{Dual Contrastive Loss}

With the feature representation $\bz_i$ and the classifier $\bt_i$ for input example $x_i$, we try to align the softmax transform of $\bt_i^T\bz_i$ with the label of $\bx_i$. Let $\bt_i^*$ denote the column of $\bt_i$, corresponding to the ground-truth label of $\bx_i$. We expect the dot product $\bt_i^{*T}\bz_i$ is maximized. Thus, we turn to learn a better representation of $\bt_i$ and $\bz_i$ with supervised signals. Here we define the dual contrastive loss to exploit the relation between different training samples, which tries to maximize $\bt_i^{*T}\bz_j$ if $\bx_j$ has same label with $\bx_i$ while minimizing $\bt_i^{*T}\bz_j$ if $\bx_j$ carries a different label with $\bx_i$.

Given an anchor $\bz_i$ originating from the input example $\bx_i$, we take $\{\bt_j^*\}_{j\in\cP_i}$ as positive samples and $\{\bt_j^*\}_{j\in\cA_i\setminus\cP_i}$ as negative samples and define the following contrastive loss:
\begin{equation}
\cL_z=\frac1N\sum_{i\in\cI}\frac{1}{|\cP_i|}\sum_{p\in\cP_i}-\log\frac{\exp(\bt_p^*\bigcdot\bz_i/\tau)}{\sum_{a\in\cA_i}\exp(\bt_a^*\bigcdot\bz_i/\tau)}
\end{equation}
where $\tau\in\mathbb{R}^+$ is the temperature factor, $\cA_i:=\cI\setminus\{i\}$ is the set of indexes of the contrastive samples and $\cP_i:=\{p\in\cA_i:y_p=y_i\}$ is the set of indexes of positive samples, and $|\cP_i|$ is the cardinality of $\cP_i$.

Similarly, given an anchor $\bt_i^*$, we can also take $\{\bz_j\}_{j\in\cP_i}$ as positive samples and $\{\bz_j\}_{j\in\cA_i\setminus\cP_i}$ as negative samples and define another contrastive loss:
\begin{equation}
\cL_\theta=\frac1N\sum_{i\in\cI}\frac{1}{|\cP_i|}\sum_{p\in\cP_i}-\log\frac{\exp(\bt_i^*\bigcdot\bz_p/\tau)}{\sum_{a\in\cA_i}\exp(\bt_i^*\bigcdot\bz_a/\tau)}
\end{equation}

The dual contrastive loss is a combination of the above two contrastive loss terms:
\begin{equation}\label{eq:dual-loss}
\mathcal{L}_{\rm Dual}=\cL_z+\cL_\theta
\end{equation}

\subsection{Joint Training \& Prediction}

To fully exploit the supervised signal, we also expect $\bt_i$ is a good classifier for $\bz_i$. Thus we use a modified version of the cross-entropy loss to maximize $\bt_i^{*T}\bz_i$ for each input example $\bx_i$:
\begin{equation}\label{eq:ce-loss}
\mathcal{L}_{\rm CE}=\frac1N\sum_{i\in\cI}-\log\frac{\exp(\bt_i^*\bigcdot\bz_i)}{\sum_{k\in\mathcal{K}}\exp(\bt_i^k\bigcdot\bz_i)}
\end{equation}

Finally, we minimize the two training objectives to train encoder $f$. The two objectives simultaneously improve the quality of the representations of the features and the classifiers. The overall loss should be:
\begin{equation}\label{eq:all-loss}
\cL_{\rm overall}=\cL_{\rm CE}+\lambda\cL_{\rm Dual}
\end{equation}
where $\lambda$ is a hyperparameter that controls the influence of the dual contrastive loss term.

In classification, we use the trained encoder $f$ to generate the feature representation $\bz_i$ and the classifier $\bt_i$ for an input sentence $\bx_i$. Here $\bt_i$ can be seen as a ``one-example'' classifier for example $\bx_i$. We regard the argmax result of $\bt_i^T\bz_i$ as the model prediction:
\begin{equation}
\hat{y}_i=\arg\max_k(\bt_i^k\bigcdot\bz_i)
\end{equation}

Figure~\ref{fig:model} illustrates the framework of the dual contrastive learning, where $e_\text{CLS}$ is the feature representation, $e_\text{POS}$ and $e_\text{NEG}$ are the classifier representations. In this concrete example, we assume that the target sample with ``positive'' class serves as the anchor, and there is a positive sample having the same class label and a negative sample having a different class label. The dual contrastive loss is designed to simultaneously attract the feature representations to the classifier representations between positive samples, and repel the feature representation to the classifier between negative samples.

\subsection{The Duality between Representations}

The contrastive loss adopts the dot product function as a measurement of the similarity between representations. This brings a dual relationship between the feature representation $\bz$ and the classifier representation $\bt$ in DualCL. A similar phenomenon appears in the relationship between the input feature and the parameter in the linear classifiers. Then, we can regard the $\bt$ as the parameter of a linear classifier such that the pre-trained encoder $f$ may generate a linear classifier for each input sample. Thus DualCL naturally learns how to generate a linear classifier for each input sample to perform the classification task.

\subsection{Theoretical Justification of DualCL}

A theoretical justification of dual contrastive learning is given in this subsection. Let $\cX=\{\bx_i\}_{i=1}^N$ and $\cY=\{y_i\}_{i=1}^N$ are the inputs and labels of $N$ training samples. The following theorem holds for dual contrastive learning.
\begin{theorem*}
Assume that there is a constant $\epsilon$ such that $p(\bx_i,y_i)\ge\epsilon$ holds for all $i\in\cI$ and $\frac{p(y_j|\bx_i)}{p(y_j)}\propto\phi(\bx_i,y_j)$:
\begin{equation}
{\rm MI}(\cX,\cY)\ge\log N-\epsilon\cL_{\rm Dual}
\end{equation}
where $\phi$ is a symmetric function that can have different definitions. In our case, $\phi(\bx_i,y_j)=(\exp(\bt_i^*\bigcdot\bz_j)+\exp(\bt_j^*\bigcdot\bz_i))/2$.
\end{theorem*}

It can be found that minimizing the dual supervised contrastive loss is equivalent to maximizing the mutual information between the inputs and labels. Please see Appendix for detailed proof.

%% file: mainresult.tex
\begin{table*}[ht]
	\centering
	\resizebox{.99\textwidth}{!}{
	\begin{tabular}{llcccccc}
		\toprule
		Model & \multicolumn{1}{c}{Method} & SST-2 & SUBJ & TREC & PC & CR & Avg. \\
		\midrule
		\multicolumn{7}{c}{10\% training data} \\
		\midrule
		\multirow{5}*{BERT}
		& CE & 
		86.05$\pm$0.24 & 93.05$\pm$0.25 & 93.29$\pm$0.22 & 91.09$\pm$0.23 & 86.58$\pm$0.29 & 90.01$\pm$0.25 \\
		& CE+SCL \cite{gunel2021supervised_contrastive_language} & 
		86.64$\pm$0.17 & 93.20$\pm$0.16 & 93.70$\pm$0.24 & 91.46$\pm$0.23 & 88.14$\pm$0.26 & 90.63$\pm$0.21 \\
		& CE+CL & 
		87.66$\pm$0.28 & 94.27$\pm$0.21 & 94.20$\pm$0.29 & 91.67$\pm$0.27 & 87.72$\pm$0.32 & 91.10$\pm$0.27 \\
		\cmidrule{2-8}
		& DualCL w/o $\mathcal{L}_{\rm dual}$ & 
		87.90$\pm$0.19 & 93.50$\pm$0.18 & 94.01$\pm$0.31 & 91.83$\pm$0.22 & 88.13$\pm$0.30 & 91.07$\pm$0.24 \\
		& DualCL & \textbf{88.40$\pm$0.20} & \textbf{94.50$\pm$0.21} & \textbf{94.93$\pm$0.23} & \textbf{92.36$\pm$0.16} & \textbf{89.01$\pm$0.28} & \textbf{91.84$\pm$0.22} \\
        \midrule
        \multirow{5}*{RoBERTa}
		& CE & 
		90.91$\pm$0.23 & 94.03$\pm$0.19 & 94.51$\pm$0.21 & 90.65$\pm$0.20 & 92.06$\pm$0.27 & 92.43$\pm$0.22  \\
		& CE+SCL \cite{gunel2021supervised_contrastive_language} & 
		91.00$\pm$0.29 & 94.37$\pm$0.30 & 94.85$\pm$0.24 & 90.82$\pm$0.20 & 92.32$\pm$0.25 & 92.67$\pm$0.26 \\
		& CE+CL & 
		91.04$\pm$0.17 & 94.47$\pm$0.19 & \textbf{95.68$\pm$0.26} & 91.90$\pm$0.14 & 92.55$\pm$0.28 & 93.13$\pm$0.21 \\
		\cmidrule{2-8}
		& DualCL w/o $\mathcal{L}_{\rm dual}$ &
		92.48$\pm$0.18 & 94.40$\pm$0.17 & 95.18$\pm$0.16 & 91.50$\pm$0.14 & 92.88$\pm$0.20 & 93.29$\pm$0.17 \\
		& DualCL & 
		\textbf{92.67$\pm$0.21} & \textbf{94.78$\pm$0.19} & 95.36$\pm$0.18 & \textbf{92.17$\pm$0.20} & \textbf{93.24$\pm$0.24} & \textbf{93.64$\pm$0.20} \\
		\toprule
		\multicolumn{7}{c}{full training data} \\
        \midrule
		\multirow{5}*{BERT}
		& CE & 
		91.19$\pm$0.23 & 96.40$\pm$0.19 & 97.21$\pm$0.20 & 95.06$\pm$0.14 & 92.09$\pm$0.24 & 94.39$\pm$0.20  \\
		& CE+SCL \cite{gunel2021supervised_contrastive_language} & 
		91.71$\pm$0.20 & 96.25$\pm$0.19 & 97.58$\pm$0.16 & 95.26$\pm$0.13 & 93.06$\pm$0.20 & 94.77$\pm$0.18 \\
		& CE+CL & 
		91.95$\pm$0.22 & 96.72$\pm$0.15 & 97.80$\pm$0.14 & 95.21$\pm$0.11 & 93.19$\pm$0.19 & 94.97$\pm$0.16 \\
		\cmidrule{2-8}
		& DualCL w/o $\mathcal{L}_{\rm dual}$ & 91.99$\pm$0.15 & 96.78$\pm$0.13 & 97.70$\pm$0.19 & 95.30$\pm$0.15 & 93.14$\pm$0.19 & 94.97$\pm$0.16 \\
		& DualCL & \textbf{92.40$\pm$0.17} & \textbf{97.20$\pm$0.17} & \textbf{98.22$\pm$0.17} & \textbf{95.56$\pm$0.14} & \textbf{93.78$\pm$0.17} & \textbf{95.43$\pm$0.16} \\
        \midrule
        \multirow{5}*{RoBERTa}
		& CE & 
		94.09$\pm$0.24 & 96.60$\pm$0.21 & 97.10$\pm$0.20 & 95.10$\pm$0.19 & 93.41$\pm$0.24 & 95.26$\pm$0.22 \\
		& CE+SCL \cite{gunel2021supervised_contrastive_language} & 
		93.65$\pm$0.20 & 96.73$\pm$0.23 & 97.18$\pm$0.19 & 95.35$\pm$0.19 & 93.60$\pm$0.17 & 95.30$\pm$0.20 \\
		& CE+CL & 
		94.33$\pm$0.21 & 97.04$\pm$0.17 & \textbf{97.52$\pm$0.15} & 95.32$\pm$0.10 & 93.49$\pm$0.25 & 95.54$\pm$0.18 \\
		\cmidrule{2-8}
		& DualCL w/o $\mathcal{L}_{\rm dual}$ & 94.41$\pm$0.23 & 96.79$\pm$0.24 & 97.10$\pm$0.25 & 95.30$\pm$0.12 & 94.01$\pm$0.25 & 95.52$\pm$0.22 \\
		& DualCL & \textbf{94.91$\pm$0.17} & \textbf{97.34$\pm$0.19} & 97.40$\pm$0.17 & \textbf{95.59$\pm$0.12} & \textbf{94.39$\pm$0.23} & \textbf{95.93$\pm$0.18} \\
  		\bottomrule
	\end{tabular}%
	}
	\caption{Accuracy on the test set. The models are trained with 10$\%$ of the training data or with full training data. We reproduce the results with the same hyperparameter configurations for all baselines for a fair comparison and report average accuracy across 10 different random seeds.}
	\label{tab:main_results}
\end{table*}

%% file: exp.tex
\section{Experiments}

\subsection{Datasets}

We conduct our experiments on following five benchmark text classification datasets. SST-2 \cite{socher2013parsing_sst2} is a sentiment classification dataset of movie reviews. SUBJ \cite{pang2004sentimental} is a review dataset with sentence labelled as subjective or objective. TREC \cite{li2002learning_trec} contains questions from six different domains, including description, entity, abbreviation, human, location and numeric. PC \cite{ganapathibhotla2008mining_procon} is a binary sentiment classification dataset that includes Pros and Cons data. CR \cite{ding2008holistic_cr} is a customer review data set and each sample is labelled as positive or negative. Table~\ref{tab:datasets} summarizes the statistics of the datasets.

\begin{table}[t]
	\centering
	\resizebox{.85\columnwidth}{!}{
	\begin{tabular}{cccccc}
		\toprule
		Dataset & \#Class & AvgLen & \#Train & \#Test & $|\text{V}|$ \\
		\midrule
		SST-2 & 2 & 17 & 7,447 & 1,821 & 15,300 \\
		\midrule
		SUBJ & 2 & 21 & 9,000 & 1,000 & 20,874 \\
		\midrule
		TREC & 6 & 9 & 5,452 & 500 & 8,751 \\
		\midrule
		PC & 2 & 7 & 32,097 & 13,759 & 9,982 \\
		\midrule
		CR & 2 & 18 & 3,394 & 376 & 5,542 \\
  		\bottomrule
	\end{tabular}%
	}
	\caption{Statistics for the five text classification datasets.}
	\label{tab:datasets}
\end{table}

\begin{figure*}[t]
    \centering
    \begin{subfigure}[t]{.33\textwidth}
        \centering
        \includegraphics[width=.95\textwidth]{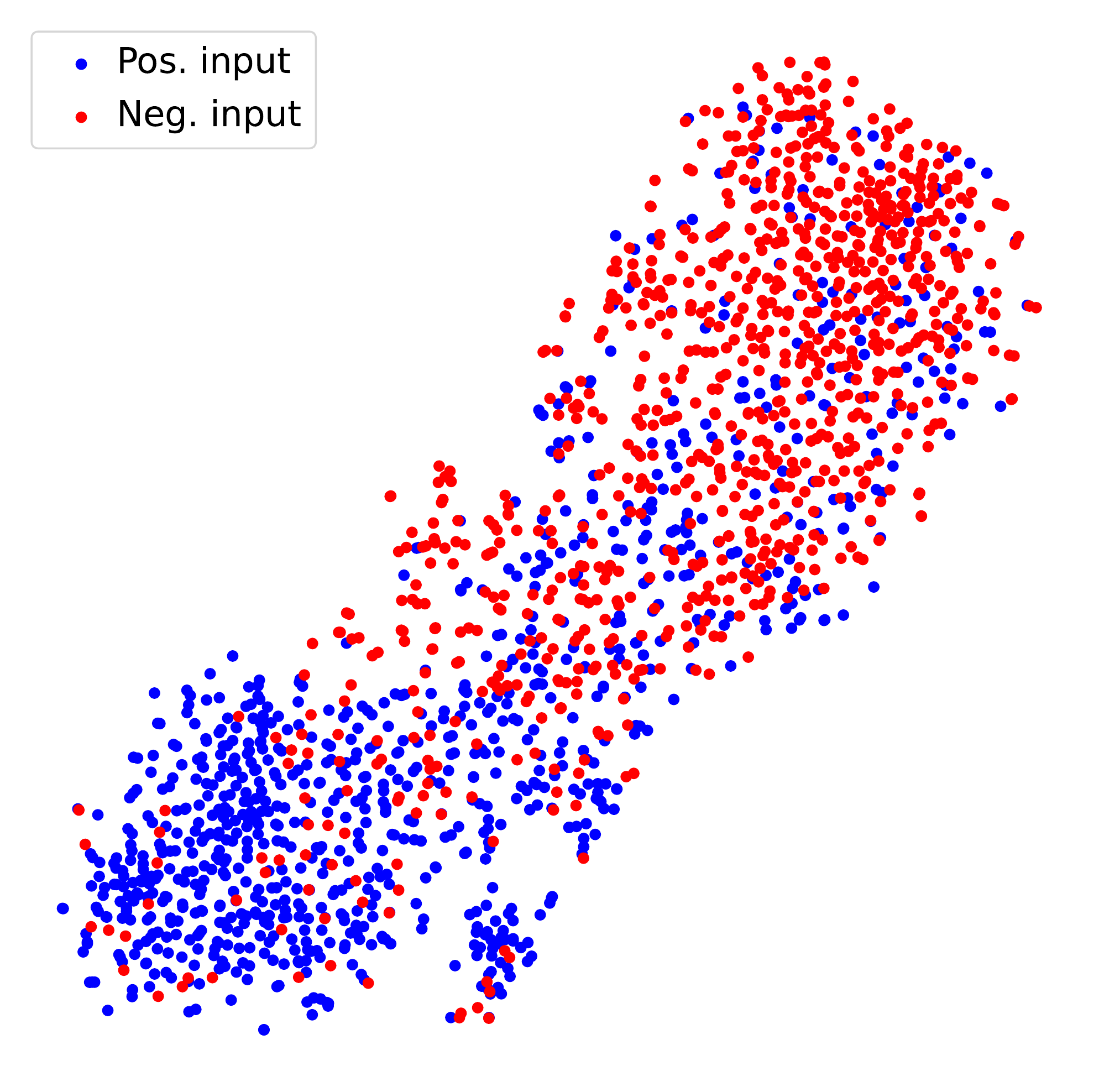}
        \caption{Cross-Entropy}
    \end{subfigure}
    \begin{subfigure}[t]{.33\textwidth}
        \centering
        \includegraphics[width=.95\textwidth]{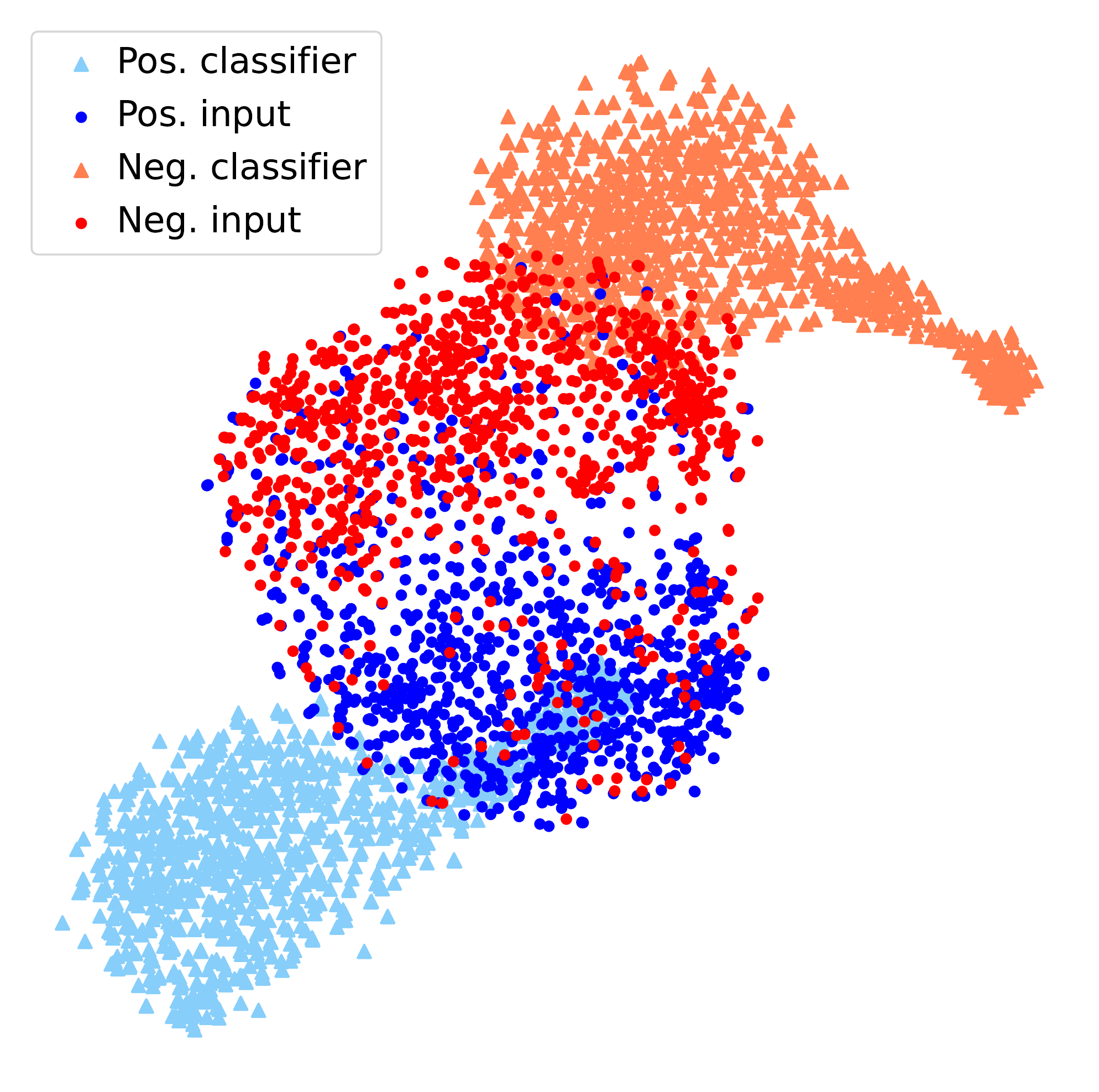}
        \caption{DualCL w/o $\cL_{\rm Dual}$}
    \end{subfigure}
    \begin{subfigure}[t]{.33\textwidth}
        \centering
        \includegraphics[width=.95\textwidth]{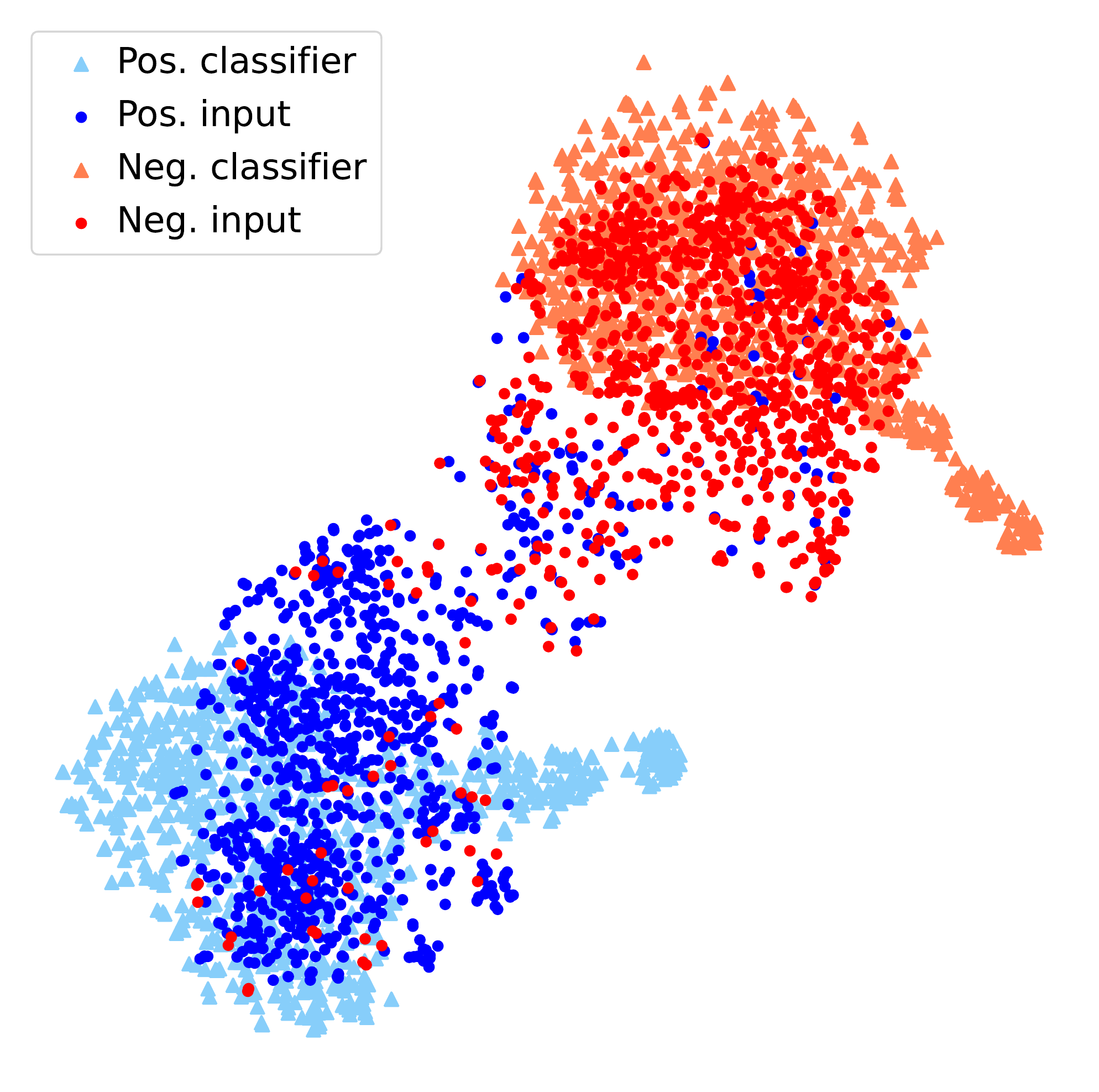}
        \caption{DualCL}
    \end{subfigure}
    \caption{The tSNE plots of the learned representations on the SST-2 dataset.}
    \label{fig:scatter}
\end{figure*}

\subsection{Implementation Details}

To adapt the input format to the BERT-family pretrained language models, for each input sentence, we list the names of all the labels as a token sequence and insert it before the input sentence, and split them with a special [SEP] token. We also insert a special [CLS] token in the front of the input sequence and append a [SEP] token at the end of the input sequence.

Both BERT and RoBERTa employ the position embeddings to make use of the order of the tokens in the input sequence, thus the class labels will be associated with the position embeddings if they are listed in a fixed order. In order to mitigate the influence of label orders, we randomly change the order of the labels before forming the input sequence in the training phase. During the test phase, the label order remains unchanged.

We use the AdamW \cite{loshchilov2018decoupled} optimizer to finetune the pretrained BERT-base-uncased and RoBERTa-base model \cite{wolf2019huggingface} with a $0.01$ weight decay. We train the model for $30$ epochs and use a linear learning rate decay from $2\times10^{-5}$ to $10^{-5}$. We set the dropout rate to $0.1$ for all layers and the batch size to 64 for all datasets. For the hyperparameters, we adopt a grid search strategy to choose the best $\lambda$ in $\{0.01, 0.05, 0.1\}$. The temperature factor $\tau$ is chosen as $0.1$. Our PyTorch implementation is available at: \url{https://github.com/hiyouga/Dual-Contrastive-Learning}.

\subsection{Experimental Results}

We compare DualCL with three supervised learning baselines: the model trained with cross-entropy loss (CE), with both the cross-entropy loss and the standard supervised contrastive loss (CE+SCL) \cite{gunel2021supervised_contrastive_language}, with both the cross-entropy loss and the self-supervised contrastive loss (CE+CL) \cite{gao2021simcse}. The results are shown in Table~\ref{tab:main_results}. 

From the results, it can be seen that DualCL with both BERT and RoBERTa encoders achieves the best classification performance in almost all settings, except on the TREC dataset where RoBERTa is employed. Compared to CE+CL with full training data, the average improvement of DualCL is 0.46\% and 0.39\% on BERT and RoBERTa, respectively. Furthermore, we observe that with 10\% training data, DualCL outperforms the CE+CL method by a larger margin, which is 0.74\% and 0.51\% higher on BERT and RoBERTa, respectively. Meanwhile, CE and CE+SCL cannot surpass the performance of DualCL. This is because the CE method neglects the relation between the samples and the CE+SCL method cannot directly learn a classifier for the classification tasks.

In addition, we find that the dual contrastive loss term helps the model to achieve better performance on all five datasets. It shows that leveraging the relations between samples helps the model to learn better representations in contrastive learning.

\subsection{Visualization}

To investigate how dual contrastive learning improves the quality of representations, we draw the tSNE plots of learned representations on the SST-2 test set. We use RoBERTa as the encoder and finetune the encoder with 25 training samples per class. We show the results of CE, DualCL without $\mathcal{L}_{\rm Dual}$ and DualCL in Figure~\ref{fig:scatter}.

In Figure~\ref{fig:scatter}, we can find that DualCL learns representations both for the input samples and the classifier associating to each sample. Comparing Figure~\ref{fig:scatter}(b) with Figure~\ref{fig:scatter}(c), we observe that the dual contrastive loss helps the model to learn more discriminative and robust representations for the input features and the classifiers, by exploiting the relation between training samples and imposing additional constraints to the model.

\subsection{Effects in Low-Resource Scenarios}

\begin{figure}[t]
    \centering
    \includegraphics[width=.495\columnwidth]{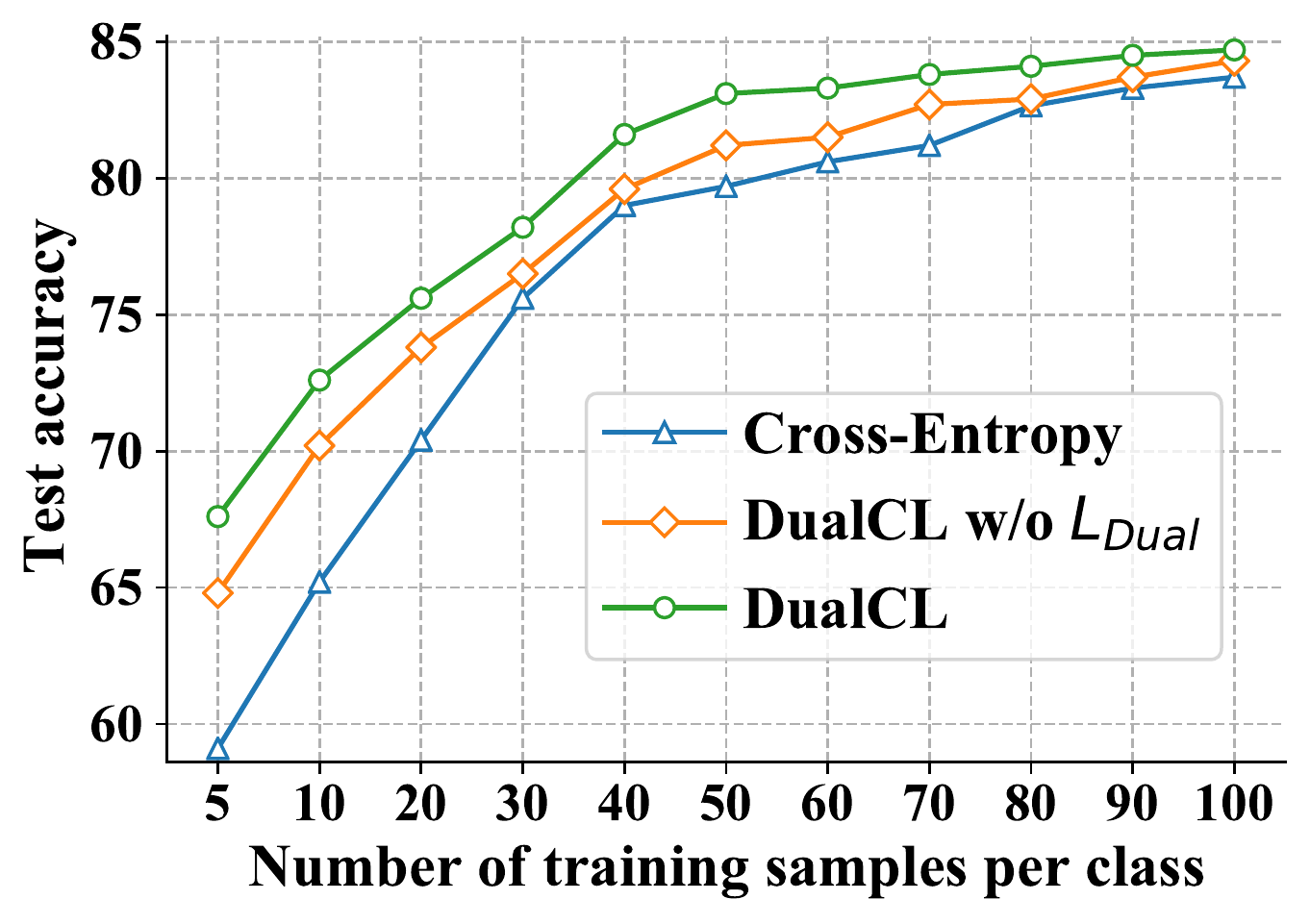}
    \includegraphics[width=.495\columnwidth]{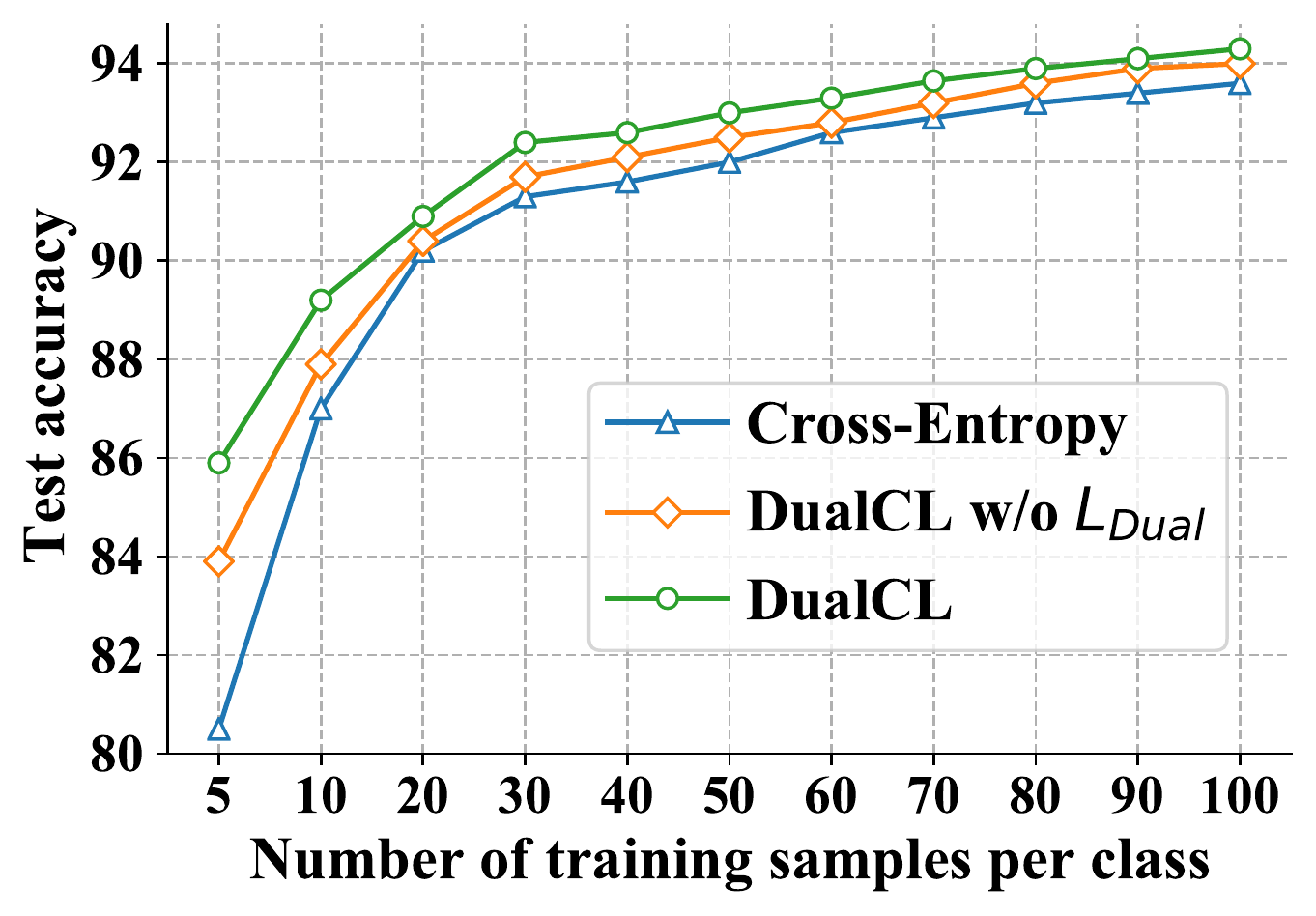}
    \caption{Test accuracy on the SST2 (left) and SUBJ (right) datasets in low-resource scenarios of DualCL with or without $\cL_{\rm Dual}$ compared with cross-entropy using BERT. The batch size here is set to $4$.}
    \label{fig:few_data}
\end{figure}

In DualCL, it uses the label-aware input representation as another view of the input samples. Thus we conjecture that the label-aware data augmentation is supposed to serve for the low-resource scenarios. To validate this, we perform experiments on the SST-2 and SUBJ datasets in low-resource scenarios. We evaluate the model performance using $N$ training samples per class, where $N\in\{5,10,20,30,40,50,60,70,80,90,100\}$. We sketch the results of BERT trained with CE, DualCL without $\mathcal{L}_{\rm Dual}$ and DualCL in Figure~\ref{fig:few_data}.

In Figure~\ref{fig:few_data}, we can see that DualCL significantly surpasses the CE method on the reduced datasets. Specifically, the improvement is up to 8.5\% on the SST2 dataset and 5.4\% on the SUBJ dataset with only 5 training samples per class. Even without the dual contrastive loss, the label-aware data augmentation can consistently improve the model's performance on the reduced dataset.

\subsection{Case Study}

\begin{figure}[t]
    \centering
    \includegraphics[width=.99\columnwidth]{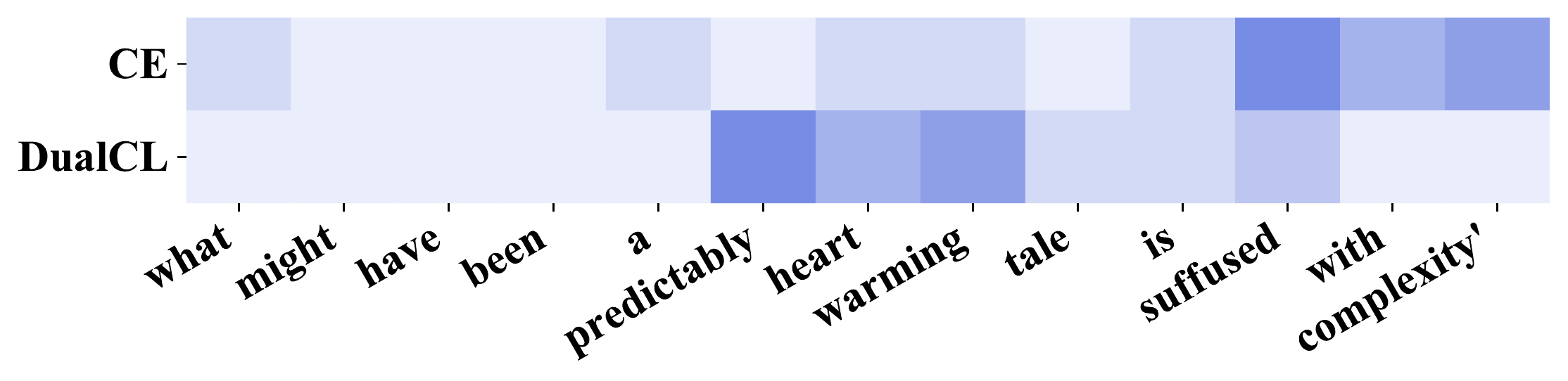}
    \includegraphics[width=.99\columnwidth]{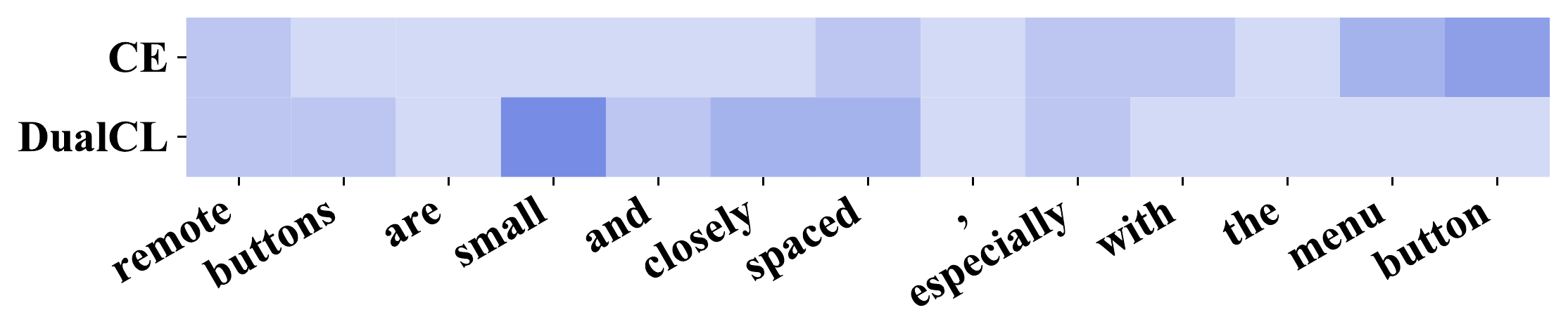}
    \caption{The visualization of attention map for CE and DualCL. The darker blue refers to higher attention scores.}
    \label{fig:case_study}
\end{figure}

To validate whether DualCL can capture informative features, we compute the attention score between the feature of [CLS] token and each word in the sentence. We firstly finetune the RoBERTa encoder on the full training set. Then we compute the $\ell_2$ distance between the features and visualize the attention map in Figure~\ref{fig:case_study}. It shows that when classifying sentiments, the captured features are different. The above example comes from the SST-2 dataset, we can see that our model pays higher attention to  ``predictably heart warming'' for the sentence expressing a ``positive'' sentiment. The below example comes from the CR dataset, we can see that our model pays higher attention to ``small'' for the sentence expressing a ``negative'' sentiment. On the contrary, the CE method fails to concentrate on these discriminative features. The results suggest that our DualCL can successfully attend to the informative keywords in the sentence.

%% file: relworks.tex
\section{Related Work}

\subsection{Text Classification}

Text classification is a typical task of categorizing texts into groups, including sentiment analysis, question answering, etc. Due to the unstructured nature of the text, extracting useful information from texts can be very time-consuming and inefficient. With the rapidly development of deep learning, neural network methods such as RNN \cite{hochreiter1997long_lstm,chung2014empirical} and CNN \cite{kim2014convolutional_cnn,zhang2015character_cnn} have been widely explored for efficiently encoding the text sequences. However, their capabilities are limited by the computational bottlenecks and the problem of long-term dependencies. Recently, large-scale pre-trained language models (PLMs) based on transformers \cite{vaswani2017attention} has emerged as the art of text modeling. Some of these auto-regressive PLMs include GPT \cite{radford2018improving_gpt} and XLNet \cite{yang2019xlnet}, auto-encoding PLMs such as BERT \cite{devlin2019bert}, RoBERTa \cite{liu2019roberta} and ALBERT \cite{lan2019albert}. The stunning performance of PLMs mainly comes from the extensive knowledge in the large scale corpus used for pretraining. 

\subsection{Contrastive Learning}

Despite the optimality of the cross-entropy in supervised learning, a large number of studies have revealed the drawbacks of the cross-entropy loss, {\em e.g.}, vulnerable to noisy labels \cite{zhang2018noisy}, poor margins \cite{elsayed2018larginmargin} and weak adversarial robustness \cite{pang2019advsample}. Inspired by the InfoNCE loss \cite{oord2018infonce}, contrastive learning \cite{hadsell2006dimensionality} has been widely used in unsupervised learning to learn good generic representations for downstream tasks. For example, \cite{he2020moco} leverages a momentum encoder to maintain a look-up dictionary for encoding the input examples. \cite{chen2020simclr} produces multiple views of the input example using data augmentations as the positive samples, and compare them to the negative samples in the datasets. \cite{gao2021simcse} similarly dropouts each sentence twice to generate positive pairs. In the supervised scenario, \cite{khosla2020supervised_contrastive} clusters the training examples by their labels to maximize the similarity of representations of training examples within the same class while minimizing ones between different classes. \cite{gunel2021supervised_contrastive_language} extends supervised contrastive learning to the natural language domain with pretrained language models. \cite{lopez2022supervised_contrastive_network} studies the network intrusion detection problem using well-designed supervised contrastive loss.

%% file: conclusion.tex
\section{Conclusion}

In this study, from the perspective of text classification tasks, we propose a dual contrastive learning approach, DualCL, for solving the supervised learning tasks. In DualCL, we simultaneously learn two kinds of representations with the pretrained language models. One is the discriminative feature of the input examples and another is a classifier for that example. We introduce the label-aware data augmentation to generate different views of the input samples, containing the feature and the classifier. Then we design a dual contrastive loss to make the classifier to be valid for the input feature. The dual contrastive loss leverages the supervised signal between the training samples to learn better representations. We validate the effectiveness of dual contrastive learning through extensive experiments. DualCL successfully achieves state-of-the-art performance on five benchmark text classification datasets. We also explain the mechanism inside dual contrastive learning by visualizing the learned representations. Finally, we find DualCL is capable of improving the model performance in low-resource datasets. Further exploration on dual contrastive learning in other supervised learning tasks, such as image classification and graph classification will be done in the future.

%% file: appendix.tex
\section*{Appendix}

\begin{theorem*}
Assume that there is a constant $\epsilon$ such that $p(\bx_i,y_i)\ge\epsilon$ holds for all $i\in\cI$ and $\frac{p(y_j|\bx_i)}{p(y_j)}\propto\phi(\bx_i,y_j)$:
\begin{equation}
{\rm MI}(\cX,\cY)\ge\log N-\epsilon\cL_{\rm Dual}
\end{equation}
where $\phi$ is a symmetric function that can have different definitions. In our case, $\phi(\bx_i,y_j)=(\exp(\bt_i^*\bigcdot\bz_j)+\exp(\bt_j^*\bigcdot\bz_i))/2$.
\end{theorem*}

\begin{proof}
Let $\phi(\bx_i, y_j)=(\psi(\bx_i, y_j)+\psi(\bx_j, y_i))/2$ and $M_i=\sum_{j=1}^{N}\frac{p(y_j|\bx_i)}{p(y_j)}$, where $\psi(\bx_i, y_j)=\exp(\bt_i^*\bigcdot\bz_j)$. We assume that $\frac{1}{|\mathcal{P}_i|}\sum_{p\in\mathcal{P}_i}\phi(\bx_i,y_p)=\phi(\bx_i,y_i)$ when $|\mathcal{P}_i|$ is sufficiently large. We have:
\begin{align*}
  {\rm MI}(\cX, \cY)
    & = \frac{1}{N} \sum_{i=1}^{N} \sum_{j=1}^{N} p(\bx_i,y_j) \log\left(\frac{p(y_j|\bx_i)}{p(y_j)}\right) \\
    & = \frac{1}{N} \sum_{i=1}^{N} \sum_{j=1}^{N} p(\bx_i,y_j) 	\left(\log\frac{p(y_j|\bx_i)}{p(y_j)M_i} + \log M_i\right) \\
    & = \frac{1}{N} p(\bx_i,y_i) \log \frac{\phi(\bx_i, y_i)}{\sum_{t=1}^{N}\phi(\bx_i, y_t)} \\
    & + \frac{1}{N} \sum_{j \neq i} p(\bx_i,y_j)\log\frac{\phi(\bx_i, y_j)}{\sum_{t=1}^{N}\phi(\bx_i, y_t)} + \log N \\
    & \geq\log N + \frac{\epsilon}{N} \sum_{i=1}^{N} \log \frac{\phi(\bx_i, y_i)}{\sum_{t=1}^{N}\phi(\bx_i, y_t)} \\
    & = \log N + \frac{\epsilon}{N} \sum_{i=1}^{N} \frac{1}{|\mathcal{P}_i|} \sum_{p\in\mathcal{P}_i} \log  \frac{\phi(\bx_i,y_p)}{\sum_{t=1}^{N}\phi(\bx_i, y_t)} \\
    & \geq \log N + \frac{\epsilon}{N} \sum_{i=1}^{N} \frac{1}{|\mathcal{P}_i|}\sum_{p\in\mathcal{P}_i} \log  \frac{\psi(\bx_i,y_p)}{\sum_{t=1}^{N}\psi(\bx_i, y_t)} \\
    & + \frac{\epsilon}{N} \sum_{i=1}^{N} \frac{1}{|\mathcal{P}_i|} \sum_{p\in\mathcal{P}_i} \log  \frac{\psi(\bx_p,y_i)}{\sum_{t=1}^{N}\psi(\bx_t, y_i)} \\
    & = \log N-\epsilon\cL_{\rm Dual}
\end{align*}
\end{proof}

This proves that the negative dual contrastive loss is a lower bound of the mutual information ${\rm MI}(\cX,\cY)$. Thus, when we minimize the dual contrastive loss, the mutual information between inputs and labels is accordingly maximized.